\documentclass{article}
\pdfoutput=1

\PassOptionsToPackage{numbers}{natbib}
    \usepackage[final,nonatbib]{neurips_2020}

\usepackage[utf8]{inputenc} 
\usepackage[T1]{fontenc}    
\usepackage{hyperref}       
\usepackage{url}            
\usepackage{booktabs}       
\usepackage{amsfonts}       
\usepackage{nicefrac}       
\usepackage{microtype}      
\usepackage{amsmath}
\usepackage[dvipsnames]{xcolor}
\usepackage{graphicx}
\usepackage{subcaption}
\usepackage[ruled,vlined]{algorithm2e}
\usepackage{amssymb}

\title{Language Through a Prism: A Spectral Approach for Multiscale Language Representations}

\makeatletter
\def\@fnsymbol#1{\ensuremath{\ifcase#1\or \dag\or \ddagger\or
   \mathsection\or \mathparagraph\or \|\or **\or \dagger\dagger
   \or \ddagger\ddagger \else\@ctrerr\fi}}
    \makeatother

\author{%
  Alex Tamkin\thanks{\texttt{atamkin@stanford.edu}} \\
  Stanford University\\
  \And
  Dan Jurafsky\\
  Stanford University\\
  \And
  Noah Goodman\\
  Stanford University\\
}

\begin{document}

\maketitle

\begin{abstract}
  Language exhibits structure at different scales, ranging from subwords to words, sentences, paragraphs, and documents. 
To what extent do deep models capture information at these scales, and can we force them to better capture structure across this hierarchy?
We approach this question by focusing on individual neurons, analyzing the behavior of their activations at different timescales. 
We show that signal processing provides a natural framework for separating structure across scales, enabling us to 1)~disentangle scale-specific information in existing embeddings and 2)~train models to learn more about particular scales. 
Concretely, we apply spectral filters to the activations of a neuron across an input, producing \emph{filtered embeddings} that perform well on part of speech tagging (word-level), dialog speech acts classification (utterance-level), or topic classification (document-level), while performing poorly on the other tasks. 
We also present a \emph{prism layer} for training models, which uses spectral filters to constrain different neurons to model structure at different scales.
Our proposed BERT + Prism model can better predict masked tokens using long-range context and produces multiscale representations that perform better at utterance- and document-level tasks.
Our methods are general and readily applicable to other domains besides language, such as images, audio, and video. 
\end{abstract}

\begin{figure}
    \centering
    \includegraphics[width=.90\linewidth,bb=0 0 843 388]{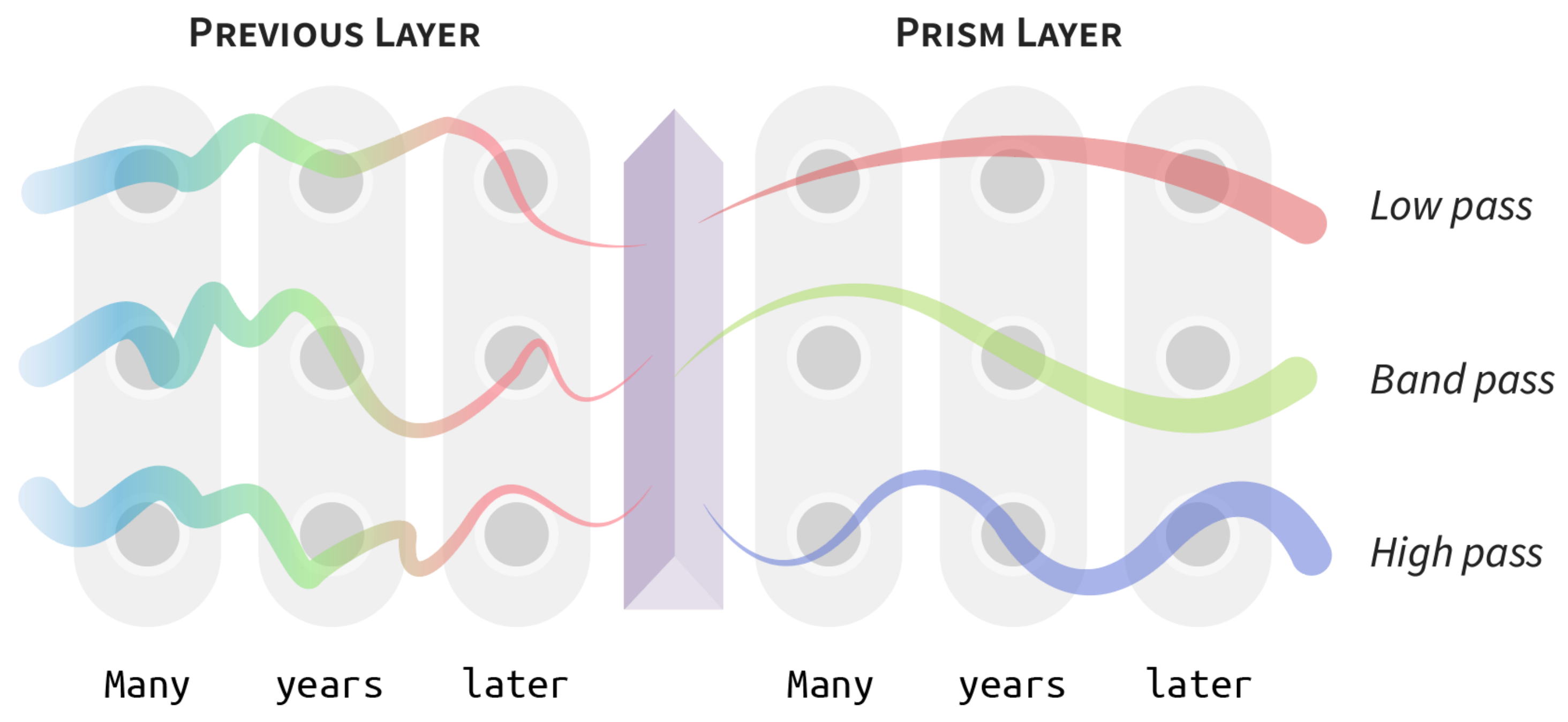}  
    \caption{\textbf{The prism layer specializes different neurons for different scales.}  First, the representations for an input are computed (left; in this case, the input is of length three). Next, a spectral filter (a low-, high-, or band-pass) is applied along the activations of each individual neuron (right). This produces neurons that are only able to represent structure at particular scales. Curved lines illustrate the scales at which neurons can change over an input.} 
    \label{fig:prism-header}
\end{figure}

\section{Introduction}

Language exhibits structure at multiple levels, ranging from morphology at the subword level \cite{nida1949morphology}, word meaning at the lexical level \cite{cruse1986lexical}, coherence and other discourse properties at the clause or sentence level \cite{kehler2002coherence, grosz1995centering, thompson1987rhetorical}, to topical and narrative structures for entire documents \cite{hearst1997texttiling, blei2003latent}. Prior work in NLP has shown how these kinds of structures can be explicitly modeled by representing individual levels of structure \cite{mikolov2013distributed, pennington2014glove, kiros2015skip, hill-etal-2016-learning-distributed, dai2015document, sinoara2019knowledge},  multiple levels of structure \cite{lin2015hierarchical, li2015hierarchical, yang2016hierarchical}, building hierarchical models that capture structure at the sentence level \cite{tai2015improved, bowman2016fast} or between sentences \cite{ji2014representation, reimers2019sentence}, and probing to discover known linguistic levels of structure \cite{conneau2018you, tenney2019bert, liu2019linguistic, hewitt2019structural}.

We propose a new method for uncovering and learning this kind of structure in representations at every scale, from word meaning to document topics, without drawing on prior linguistic models of specific structural levels like "sentence" or "clause." To do so, we employ tools from spectral analysis, widely used in signal processing and other fields \cite{oppenheim1999discrete} to separate and control information at different timescales. Intuitively, any sequence of values, such as a neuron's activations across input tokens, can be represented as a weighted sum of cosine waves with different frequencies. The weight for a particular frequency indicates the amount of structure in the sequence at that scale: weight on higher frequencies indicates faster changes in the neuron's activation from token to token, while weight on lower frequencies indicates activations that shift more gradually across an input. By removing certain frequencies, called \emph{spectral filtering}, we can remove information about variation at particular scales. See Figure \ref{fig:filters} for a visualization.

In this work, we apply spectral filters to the activations of individual neurons in BERT \cite{devlin2018bert}, a popular deep NLP model. This enables us to separate information in model representations that changes at different rates across the input---for example, part of speech changes on a word-to-word basis, while topical changes are much more gradual. Concretely, we contribute: 

\begin{enumerate}
    \item \textbf{A principled framework} based on spectral analysis for describing structure at multiple scales in deep representations. While we consider applications to NLP models, this is a general framework that could extend to other models with representations arranged in spatial or temporal structure. (Section~\ref{sec:filters})
    \item \textbf{A technique}, \emph{spectral filtering}, for extracting scale-specific information from language representations. We show how low-pass filters can alter representations to only perform well on topic classification (document-level), while band-pass and high-pass filters do the same for dialog acts classification (utterance-level) and part of speech tagging (word-level). (Section~\ref{sec:probing})
    \item \textbf{A new model component}, the \emph{prism layer}, which specializes neurons in a model for particular scales of structure. After training with a prism layer, our model is more sensitive to long-range interactions between tokens and produces individual representations that perform comparably or better than BERT's across tasks at different scales. (Section~\ref{sec:prism})
\end{enumerate}
\section{Spectral filtering of contextual word representations}
\label{sec:filters}

This section provides some background on the spectral analysis tools we use and describes how we apply them to deep language representations.

\subsection{Background: The discrete cosine transform and spectral filters}
\label{subsec:background}

In order to perform operations in the frequency domain of a sequence, we first need to obtain a representation of the input in the frequency domain. This is the role of a \emph{spectral transform}. The spectral transform we use in this work is the discrete cosine transform (DCT\footnote{More precisely, this transform is known as the DCT-II}) \cite{rao2014discrete}, a widespread tool used in audio coding, texture analysis, image classification, and compression \cite{zuo2014boundary,rao2014discrete}. The DCT represents a real-valued sequence of points as a same-length sequence of weights over cosine functions of different frequencies. Formally, for a real-valued sequence $\{x^{(0)} \ldots x^{(N-1)}\}$ its DCT (the weights for each frequency) is obtained by
\begin{align}
\label{eq:dct}
f^{(k)} = 
\sum_{n=0}^{N-1}
x^{(n)}\cos 
\left [ 
    \frac{\pi}{N}
    \left(
        n + \frac 1 2
    \right) k
\right ]
\qquad
k = 0, \ldots , N-1
\end{align}

Intuitively, the DCT computes the similarity of a signal and cosine waves of different frequencies by taking the dot product between them. These dot products constitute the coefficients of the signal in the frequency domain. The DCT is closely related to the discrete Fourier transform (DFT). We use the DCT here because it is a real-to-real function (the DFT is complex-to-complex), is widely used in practice, and can often produce fewer artifacts than the DFT when filtering \cite{rao2014discrete, bovik2009essential}.

\begin{figure}
    \centering
    \includegraphics[width=.95\linewidth,bb=0 0 840 214]{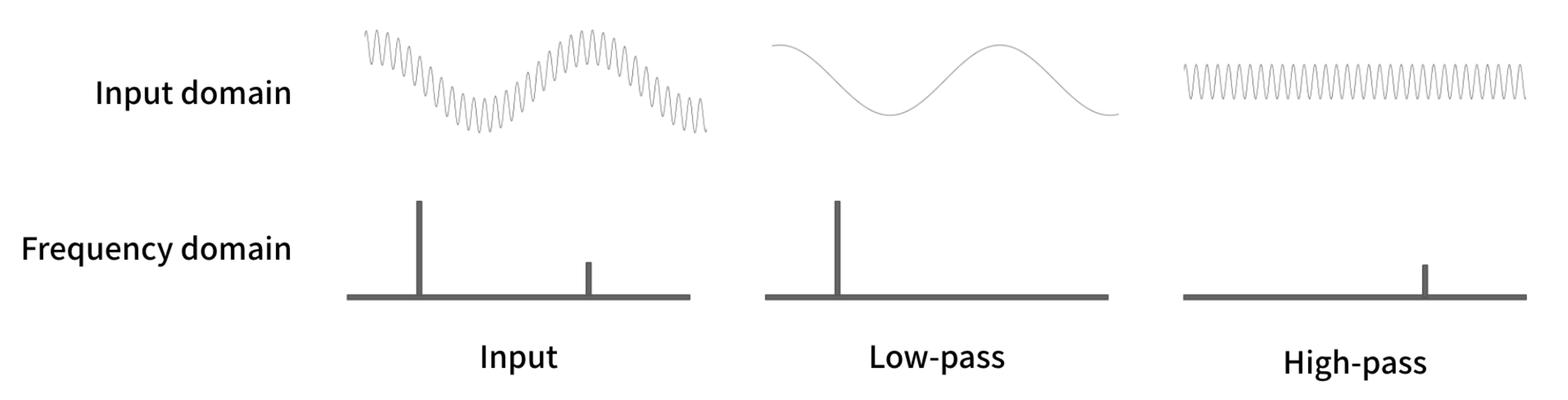}  
    \caption{\textbf{A visual depiction of spectral filters and their effects in the input and frequency domain.} The input domain shows a sequence of values (e.g., the activation of a neuron across input tokens). The frequency domain shows the weight on the cosine waves which sum to produce the curve in the input domain. Low-pass filters only allow low frequencies to pass through, producing a smoothed input. High-pass filters only allow high frequencies and produce a locally-normalized input. Band-pass filters (not shown) are compositions of low- and high-pass filters.}
    \label{fig:filters}
\end{figure}

The DCT of a sequence enables straightforward manipulation of structure at different scales in a sequence. For example, one can remove components above some threshold frequency $k_{\mathrm{thresh}}$ by setting $f_k \leftarrow 0$ for all $k > k_{\mathrm{thresh}}$, then applying the inverse DCT (IDCT) to return the signal to the original domain \cite{ahmed1974discrete}. This is known as a \emph{low-pass filter}, and returns a smoothed, same-length version of the original input, removing shorter-term fluctuations. The inverse operation can be performed to achieve a \emph{high-pass filter}, which returns a signal where each term is locally normalized with respect to its neighbors, neutralizing longer-term trends. Composing these two operations yields a \emph{band-pass filter}, as only a band of frequencies is allowed to pass through the filter. See Figure \ref{fig:filters} for a visual depiction.\footnote{Fully zeroing out frequencies (a \emph{brick wall filter}) can produce artifacts after performing the IDCT, motivating the use of smoother attenuation functions \cite{butterworth1930theory, takahasi1951ladder}, which reduce artifacts in exchange for allowing less-than-full attenuation of frequencies outside the desired band. However, for simplicity, we use brick wall filters in this work, leaving study of other filters, as well as other spectral tools like wavelets \cite{mallat1999wavelet}, for future work.}

\subsection{Applying the DCT to contextual word representations}

How do we apply the DCT to deep language representations? A common feature of modern NLP models is \emph{contextual word representations}, a sequence of vectors created by processing a sequence of tokens (e.g., words or subword units). These representations are produced by a wide range of modern NLP architectures, including Transformer-based \cite{vaswani2017attention} models like BERT \cite{devlin2018bert} and GPT-2 \cite{radford2019language}, as well as LSTM-based \cite{hochreiter1997long} models such as ELMo \cite{peters2018deep}. 

Assume we are given a sequence of contextual word representations $v_{0}, \ldots, v_{N-1}$. The core technique we propose is to apply the DCT to a slice of these representations \emph{along a single neuron}: $v_0[i], \ldots, v_{N-1}[i]$. We refer to the transformed sequence in the frequency domain $f_0[i], \ldots, f_{N-1}[i]$ as the spectrum of the $i$th neuron. $f_0[i]$ is the lowest frequency term, corresponding to the average value of $v_0[i], \ldots, v_{N-1}[i]$, while $f_{N-1}[i]$ is the highest frequency term. We can then implement any of the filters from Section \ref{subsec:background} by zeroing out the appropriate values in the spectrum, and then applying the IDCT to return the sequence to the original domain. In practice, external libraries make this quite simple: we show a three-line implementation of a low-pass filter in Figure \ref{subfig:code}.
\section{The relationship between spectral frequencies and linguistic phenomena}
\label{sec:probing}

We have seen how to apply spectral filters to the hidden states of deep NLP models. In this section, we explore how these spectral filters can be used to separate out phenomena at different scales in contextual word representations.

\subsection{Disentangling scale-specific information in representations}
Contextual word representations have been shown to not only encode the meaning of tokens in context \cite{peters2018deep}, but also a wider range of linguistic phenomena such as semantic roles, entity types, constituent labels, relations between entities, and coreference  \cite{tenney2019you}. This suggests that these representations may already be encoding information about multiple scales ranging from the (sub)word itself to its containing phrase, clause, sentence, paragraph and perhaps the document as a whole. 
In this work, we consider whether these phenomena can be separated out at the level of \emph{individual neurons} by using spectral filters to tease apart structure at different scales in a neuron's activations across an input.

To investigate, we observe how the choice of spectral filter affects the ability of a classifier to perform tasks at different scales using the filtered representations. Each spectral filter is determined by a corresponding \emph{spectral band}: the range of frequencies that is used for the low-, high-, or band-pass. We seek to choose bands corresponding to different scales. However, the scale of a particular frequency is revealed by its period: the number of tokens it takes to complete a full cycle. For example, from Equation \ref{eq:dct} we see that index $8$ of the DCT has a frequency of $2*8 = 16$, and thus for inputs of size $512$ has a period of $512 / 16 = 32$ tokens. 

\begin{figure}
  \centering
  \begin{subtable}{.59\textwidth}
    \centering
    \begin{tabular}{llrr}
      \toprule
      Filter  & Ex. Scale & Period (toks)  & DCT index \\
      \midrule
         \textsc{high}       & Word & 1--2 & 130--511  \\
         \textsc{mid-high}  & Clause & 2--8 & 34--129  \\
         \textsc{mid}        & Sentence & 8--32 & 9--33  \\
         \textsc{mid-low}   & Paragraph &  32--256 & 2--8 \\
         \textsc{low}        & Document & 256--$\infty$ & 0--1 \\
      \bottomrule
  \end{tabular}
  \caption{\textbf{The spectral filters we consider in this work}, along with their periods, spectral bands (the indices in the DCT), and example linguistic phenomena at that scale. The period of a cosine wave for a DCT index is the approximate number of tokens it takes for the wave to complete a cycle.}
  \label{table:bands}
  \end{subtable}
  \qquad
  \begin{subfigure}{.31\textwidth}
    \centering
    \includegraphics[width=.90\linewidth,bb= 0 0 574 216]{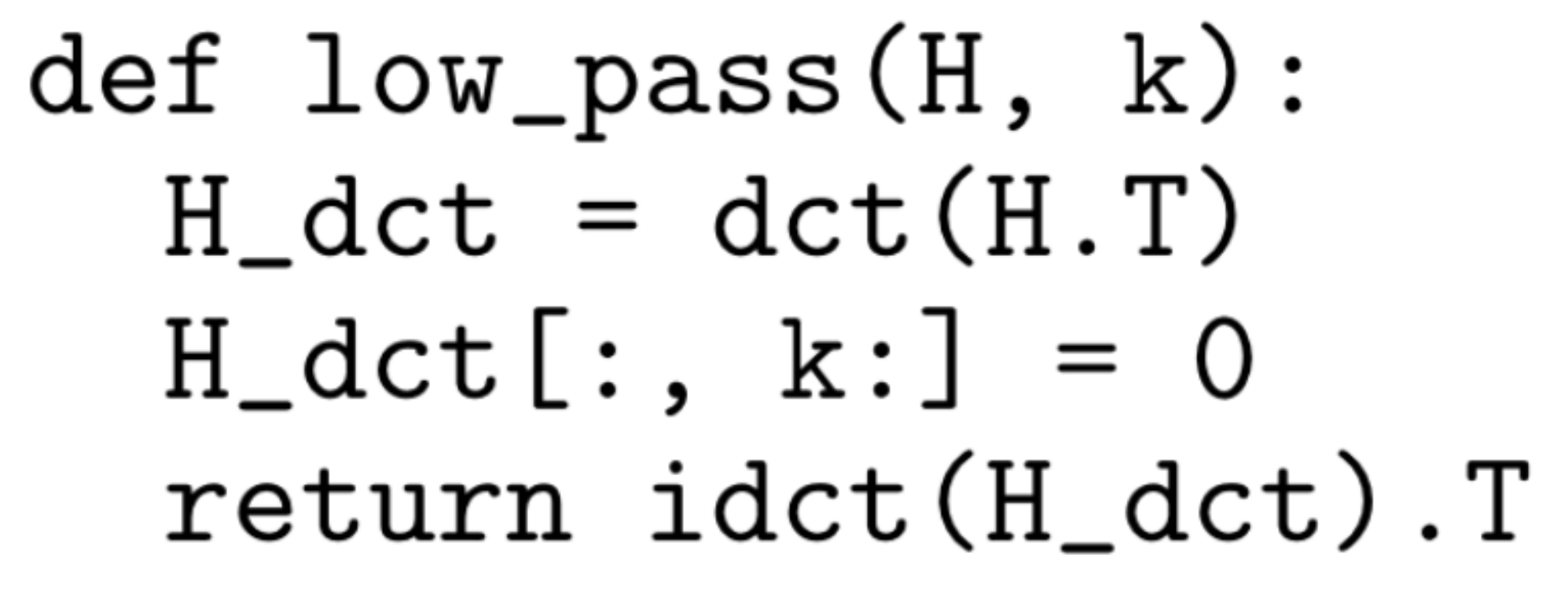}  
      \caption{\textbf{Spectral filters are simple to incorporate into existing models.} Python-style code for a low-pass filter over representations. Input \texttt{H} is a list of representations for each input token, while \texttt{k} is the low-pass threshold frequency. \texttt{T} is the transpose operator. We use a PyTorch library to compute the (I)DCT.}
    \label{subfig:code}
  \end{subfigure}
  \label{fig:decomp}
  \caption{}
\end{figure}

In this work, we divide the frequency spectrum into five bands, chosen reflect the inductive bias that linguistic units at one scale are composed of multiple units from the scale below (e.g. several words compose a phrase). Thus, we allocate bands such that for each band, the periods of the frequencies in the next higher band decay by a fixed amount. This produces five bands (\textsc{low, mid-low, mid, mid-high,} and \textsc{high}) with a diverse range of scales, as shown in Table \ref{table:bands}.\footnote{We use these five separate bands in part for instructive purposes; however, in practice, one might wish to smoothly change the endpoints of the spectral band across neurons. One could also specifically choose bands for a task based on their corresponding periods to include or exclude particular scales of interest.} See the Appendix for more details on band allocation and discretization.

\subsection{Probing bandpassed representations for linguistic information}

We evaluate the content of these filtered representations through probing experiments \cite{alain2016understanding, ettinger2016probing, shi2016does}. For each dataset below, we encode each training example with a fixed, pretrained BERT-Base cased model \cite{devlin2018bert}. This produces a series of 768-dimensional contextual word representations. We then apply a spectral filter along each dimension and train a softmax classifier to perform a particular task using each filtered representation. We examine three English-language tasks, involving classification of word-, utterance-, and document-level phenomena, providing a natural testbed for investigating the content of these representations: 
\begin{enumerate}
  \item \textbf{Part of speech tagging (word-level):} We use the Penn Treebank dataset \cite{marcus1993building}. The task is to predict the part of speech (e.g. \textsc{past tense verb}, \textsc{wh-pronoun}, \textsc{cardinal number}) from the given token representation.
  \item \textbf{Dialog speech act classification (utterance-level):} We use the Switchboard Dialog Speech Acts corpus \cite{Jurafsky-etal:1997, Shriberg-etal:1998, Stolcke-etal:2000}.\footnote{We use the preprocessing library from \url{https://github.com/cgpotts/swda}} The task is to predict the dialog speech act (e.g. \textsc{apology}, \textsc{hedge}, \textsc{appreciation}) of the utterance containing the given token representation.
  \item \textbf{Topic classification (document-level):} We use the 20 Newsgroups dataset \cite{lang1995newsweeder}. The task is to predict the topic (newsgroup; e.g. \textsc{sci.space}, \textsc{comp.graphics}, \textsc{rec.autos}) of the document containing the given token representation.
\end{enumerate}

We train our probing models for a maximum of 30 epochs, using the Adam optimizer \cite{kingma2014adam} with default parameters. We use early stopping with a patience of one, decaying the learning rate by a factor of 2 when successive epochs do not produce a decrease in validation loss. To compare against the masked language modeling (MLM) task, which was the original target task\footnote{We do not consider the next sentence prediction task (NSP) \cite{devlin2018bert}. While it was also used for BERT pretraining, we discard the \texttt{[CLS]} tokens, which are used to predict the NSP label.} 
for these representations \cite{devlin2018bert}, we also train an MLM probe for three epochs on the WikiText-103 dataset \cite{merity2016pointer}.

\begin{figure}
  \centering
  \includegraphics[width=.56\linewidth,bb=0 0 593 593]{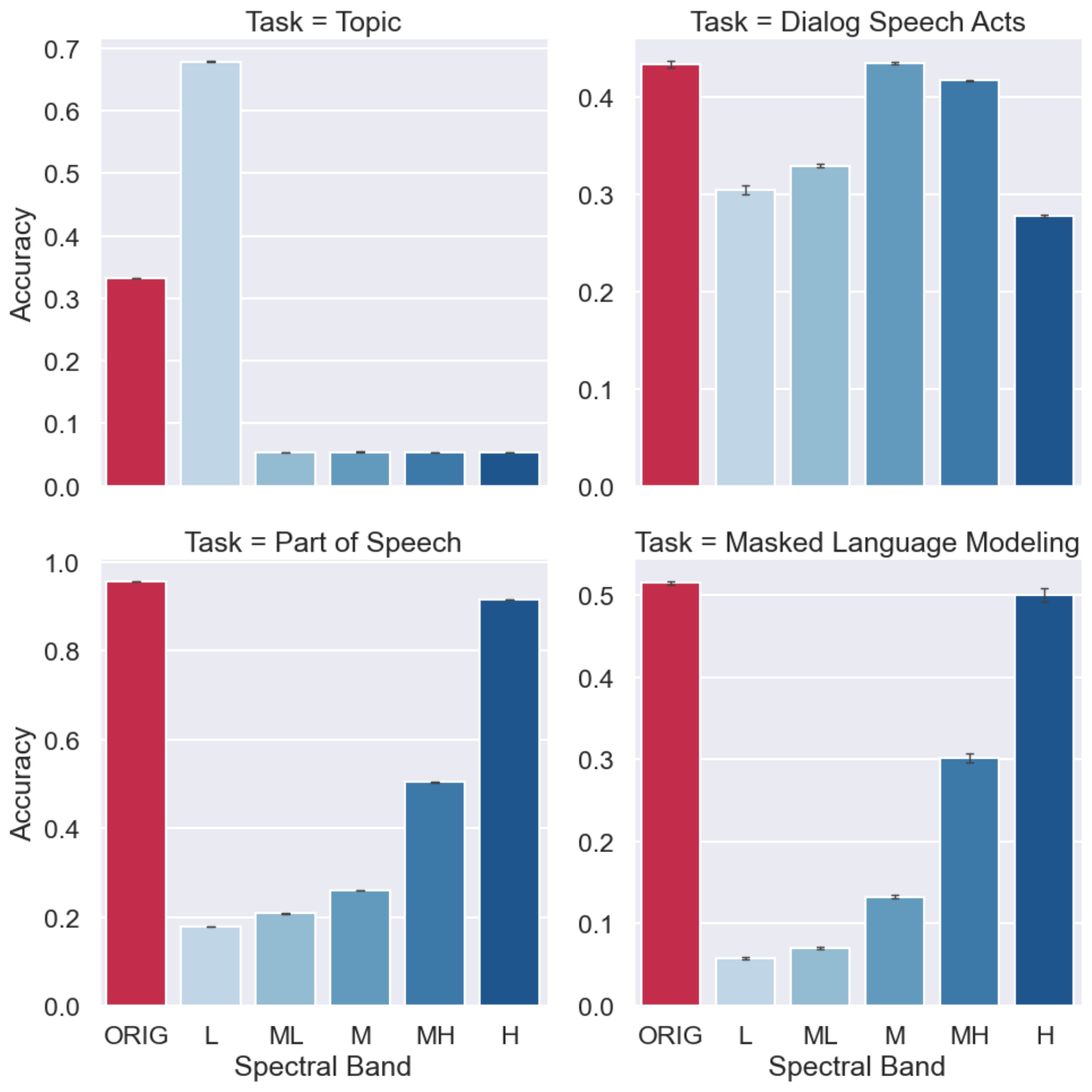} 
\caption{\textbf{Different spectral filters extract information useful for tasks at different scales.} Probing accuracy for different tasks and band-passes. A low-pass filter produces representations that yield highest probing accuracy on topic classification, while high-passed representations have highest probing accuracy for part of speech tagging. Meanwhile, band-passing the middle frequencies is most useful for dialog speech act probing. ``ORIG'' refers to the performance of the original token representations. Error bars show standard deviations over three probing runs.}
\label{fig:probing}
\end{figure}

As Figure \ref{fig:probing} shows, different spectral filters indeed produce representations specialized for the expected task. The highest probing accuracy for part of speech tagging occurs when extracting the \textsc{high} band, aligning with the fact that this is a word-level task. However, the highest frequency spectral band still performs worse than the original representations, suggesting that lower frequency information is sometimes necessary for this task (e.g. for parts of speech correlated over several tokens, such as strings of numbers or lists of nouns). By contrast, topic-classification performs best with information from the \textsc{low} band, aligning with the fact that it is a document-level phenomenon. Interestingly, the accuracy for the \textsc{low} band is substantially higher than for the original representations, suggesting that higher frequency variation present in the original representations may be harmful for that task. Meanwhile, probing for dialog speech acts, a classification task over utterances, is most successful at the \textsc{mid} band, with performance comparable to that of the original representations. The probing results for masked language modeling are most similar to part of speech tagging, underscoring the degree to which MLM is a local task.

These results demonstrate that spectral filters are effective tools for separating multiscale linguistic phenomena in contextual word representations.
\section{Using spectral filters during training}
\label{sec:prism}

In the previous section, we saw how spectral filters can be used to isolate information about linguistic phenomena at different scales in an existing model's representations. However, this observed structure arose naturally from BERT's masked language modeling task, which we saw is a relatively local task. In this section, we will show how spectral filters can be used \emph{during training} to produce  multiscale representations with improved performance on mid-scale and global tasks despite being trained with masked language modeling.

\subsection{The prism layer}
\label{subsec:prism}
In BERT, the information for the different tasks discussed above may be distributed across all neurons, rather than specialized in particular ones. 
Spectral filters, however, provide a natural way to force BERT to use different neurons for information about different scales. The resulting multiscale representations may then be better suited for a broader range of tasks than the original BERT representations.

To accomplish this, we take a given hidden state in BERT and divide the units evenly into five \emph{sectors}.\footnote{We distribute the $768 \pmod 5 = 3$ remaining units to the \textsc{low}, \textsc{low-mid}, and \textsc{mid} bands.} To each sector, we then apply a different band-pass from Table \ref{table:bands}. We call these additional computations a \emph{prism layer}, as they separate out the different frequencies in a layer's representations. See Figure \ref{fig:prism-header} for an illustration. In our main experiments, we apply one prism layer after the last BERT layer. See the Appendix for an investigation of placing prism layers after each BERT layer.

We then train our pretrained BERT model with the prism layer on the masked language modeling task; this is so the model can adjust to the new constraints imposed upon it and learn to allocate information at particular frequencies to the correct sectors. We use an external PyTorch library for computing and backpropagating through the DCT and IDCT. \footnote{\url{https://github.com/zh217/torch-dct}} We train on the WikiText-103 dataset \cite{merity2016pointer} for 50k steps at a batch size of 8 with default parameters for Adam. To allow for fair comparisons between our model and BERT, we also further train an unmodified pretrained BERT model using this same data and procedure (see the Appendix for an ablation of this step).

\subsection{Results}

We now compare the probing performance of the vanilla BERT model with the BERT model trained with our prism layer, shown in Table \ref{table:prism-probing}. The BERT model with the prism layer performs considerably better than BERT on topic (+18.8\%) and dialog speech act (+6.9\%) classification while maintaining high accuracy on part of speech tagging (-1.5\%). These results demonstrate that the prism layer has enabled BERT to produce more general-purpose representations that capture phenomena across scales.

\begin{table}
  \centering
\caption{\textbf{Training with a prism layer produces multiscale representations that perform comparably or better than BERT across different tasks.} Probing accuracy and standard deviation (3 trials) for different tasks on the final-layer BERT and BERT + Prism representations.}
\vskip 0.5cm
  \begin{tabular}{llrr}
    \toprule
    Task & Model &  Accuracy (\%) & S.D. (\%) \\
    \midrule
    Topic classification & BERT &    32.21 &  0.08 \\
    & \textbf{BERT + Prism} &    \textbf{51.01} &  0.14 \\
    \midrule
    Dialog speech acts & BERT &    47.09 &  0.33 \\
                      & \textbf{BERT + Prism} &  \textbf{54.02} &  0.61 \\
                      \midrule
    Part of speech & \textbf{BERT} &    \textbf{95.86} &  0.02 \\
                  & BERT + Prism &    94.41 &  0.02 \\
    \bottomrule
  \end{tabular}
  \label{table:prism-probing}
\end{table}

\subsection{Sensitivity to distant tokens}

The multiscale representations produced by the prism layer are used by the model to perform the masked language modeling (MLM) objective. Since these representations contain information at different scales, this provides an inductive bias for the model to rely on both long-range and short-range information when performing the MLM task. To show this quantitatively, we consider an MLM problem where one hundred consecutive tokens in the middle of the input are masked. The model's loss on these tokens reflects the model's ability to rely on distant information to predict tokens without local context. 

\begin{figure}
    \centering
    \includegraphics[width=.95\columnwidth,bb=0 0 843 272]{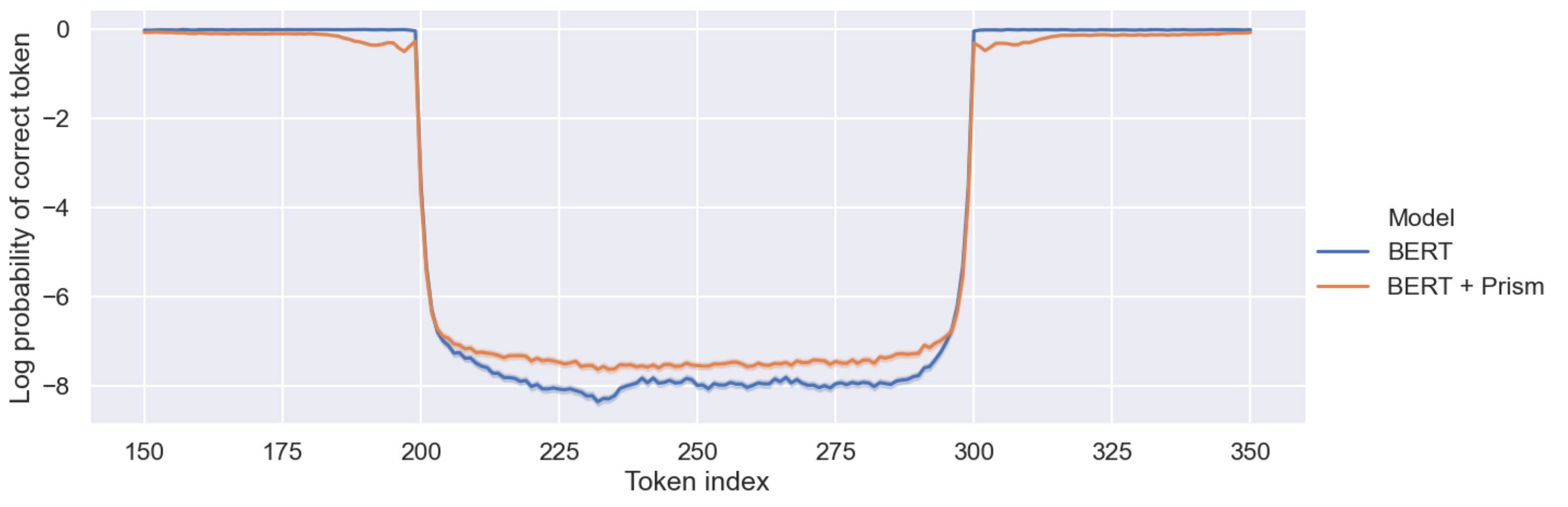}
    \caption{\textbf{Training with a prism layer significantly improves prediction of masked tokens without local context (note the log scale).} Average log probability of correct token for different indices (N=1600). Indices between 200 and 300 are replaced with a \texttt{[MASK]} token in the input, requiring the model to use long-range context to generate a probability distribution for the missing token. The higher log probabilities in the masked region for the BERT + Prism model suggest the prism layer makes the model more sensitive to long-range dependencies. Shaded regions are 95\% bootstrap CIs (generally too small to see without magnification).}
    \label{fig:cliff}
\end{figure}

We plot the average log probability of the correct token in Figure \ref{fig:cliff}, for both the BERT model with the prism layer, as well as a BERT model trained on WikiText-103 for the same number of steps. As expected, no model can precisely guess the missing tokens with perfect accuracy. But we do see a noticeable difference between the probabilities assigned to the correct token by the BERT models with and without the prism layer (note the log scale). This indicates that the BERT + Prism model is a better \emph{long range} language model, using context to predict distant tokens.

Another interesting phenomenon in the graph is the dips in log probability exhibited by the BERT + Prism model adjacent to the redacted text, indicating dependence on (redacted) distant context. No such dip exists for the original BERT model, indicating that it solves the MLM task in a very local way.
These results suggest that the prism layer is a useful tool for encouraging modeling of long-range dependencies in Transformer models \cite{trinh2018learning,hochreiter1997long}.

\section{Related work}
\label{sec:related}

Our work connects with several streams of research investigating multiscale structure in natural language and our models of it. Prior work has studied the extent of this structure at different scales in linguistic corpora, using tools ranging from random walk models and power spectra \cite{ebeling1995long, pavlov2001scaling} to entropy and mutual information \cite{ebeling1994entropy}. To model this structure, researchers have conducted multiresolution analyses of text corpora by applying diffusion wavelets to term-document corpora \cite{coifman2006diffusion}, multinomial topic distributions \cite{wang2009multiscale}, and term-term cooccurrence graphs \cite{jain2014short}. Concerning deep learning, several works have considered the challenges of modeling different scales in distributed representations of words \cite{mikolov2013distributed, sarkar2019scale} and of capturing long-term dependencies in recurrent neural networks \cite{hochreiter1997long, hochreiter2001gradient}. Other work conducts analytic studies of models that illuminate their scale-awareness, including the sensitivity of LSTM language models to relationships at different scales \cite{khandelwal2018sharp} and the attention patterns of Transformer models \cite{clark2019does}. In conversation with this literature, our work provides a principled way of understanding multiscale structure in the representations of deep models, illuminating the linguistic phenomena captured at each of these scales and enabling the construction of scale-specific representations for downstream purposes.

In concert with these analyses, a large body of work has attempted to leverage the expressive capability of distributed representations to improve modeling at particular scales. For example, several works introduce different kinds of architectural modifications to recurrent neural networks in order to encourage learning hierarchical structure, especially long-term structure, including via updating hidden states at different intervals \cite{koutnik2014clockwork}, multilayered models \cite{el1996hierarchical, chung2016hierarchical}, incorporating tree structures \cite{tai2015improved} or syntactic parsing \cite{bowman2016fast}, introducing residual connections \cite{yue2018residual}, adding auxiliary losses \cite{trinh2018learning}, and discretizing ordinary differential equations \cite{chang2019antisymmetricrnn}. In addition, certain works explicitly focus on creating high-quality representations at particular scales, including the word-level \cite{mikolov2013distributed, pennington2014glove}, sentence-level \cite{kiros2015skip, hill-etal-2016-learning-distributed, arora2016simple, reimers2019sentence}, paragraph-level \cite{dai2015document} and document-level \cite{sinoara2019knowledge}. Perhaps most similar to our work is a stream of work incorporating the Fourier basis into recurrent architectures \cite{zhang2000forenet, zhang2018learning}. However, while these works focus on speeding up training or improving gradient flow in RNNs, our approach is architecture-agnostic, provided the model produces contextual word representations, and can be used to understand or improve specific scales of interest in the model's representations. Another piece of related work is Ordered Neurons \cite{shen2018ordered}, which enforces an update hierarchy in the latent state of an RNN to capture tree-like structure in an input (e.g., syntax trees). By comparison, our approach generalizes beyond RNN or autoregressive architectures and can capture both syntactic structure like part of speech as well as longer-range multiscale phenomena like dialog speech acts and topic where tree structures may not be as appropriate.
 
Finally, our work is related to spectral approaches in audio \cite{abdel2012applying, hinton2012deep}, where it is naturally suited as an input representation, as well as computer vision, where the Fast Fourier Transform \cite{nussbaumer1981fast} and the Discrete Cosine Transform \cite{ahmed1974discrete} have been used to speed up the training of convolutional neural networks \cite{mathieu2013fast}, generate filters for scene classification \cite{khan2017scene}, and compress convolutional models \cite{cheng2015exploration}. Concerning scales, the authors of StyleGAN \cite{karras2019style} investigate how different layers in their model are responsible for phenomena at different scales, such as pose, lighting, face shape, and finer facial features. Most related to our work is a line of research that improves training by using spectral filters to replace downsampling operations in convolutional models \cite{rippel2015spectral} as well as improving optimization speed and generalization by removing low-magnitude \cite{khan2019regularization} or high-frequency \cite{rippel2015spectral} spectral coefficients. We also explore attenuation of different frequency coefficients, but in an NLP context to improve modeling of long-range dependencies, and further use spectral techniques to understand, control, and improve modeling at different scales.
\section{Conclusion}

In this work, we demonstrate how techniques from spectral analysis provide a principled and effective framework for separating multiscale phenomena in deep language representations. We first demonstrate how spectral filters can be used to separate information at different scales in BERT representations. We use this technique to produce scale-disentangled representations that perform well at either part of speech tagging, dialog acts classification, or topic classification, while performing poorly on the other two tasks. We also show how to create multiscale representations by training with a prism layer, which forces different neurons to capture information about different scales. The representations produced by the resulting model enable comparable or higher performance across the three tasks than vanilla BERT representations. We also show that training with a prism layer increases the model's sensitivity to long-range context, as measured by a masked language modeling task. These results demonstrate that spectral techniques are a powerful set of tools for uncovering and modeling multiscale phenomena in deep NLP models.

Our work provides multiple avenues for further study. For interpretability researchers, these tools could enable better understanding of knowledge and information processing at different scales in neural models across different tasks, inputs, and layers. For researchers of linguistic change, this method may enable better tracking of topics over time or facilitate the removal of extraneous information (e.g., topic) when targeting a linguistic phenomenon at a different scale. Finally, we also see promise for improving NLP models during training in a broader range of applications and architectures. More generally, we emphasize that our method is domain agnostic: it needs only a collection of representations with some kind of geometric (e.g., spatial or temporal) structure---thus, we are optimistic about the potential for further applications of these techniques on the hidden states of computer vision, time series, and reinforcement learning models, among others..

\section*{Broader Impact}

The spectral tools we provide in this paper are applicable to a wide range of neural network models and possible end uses. While this makes it difficult to speak with confidence about broader impacts of the research, we briefly discuss a few potential use cases. Scale isolation enables users to remove information about particular kinds of structure inside existing representations. This could be useful for interpretability or fairness research, as well as computational social scientists who wish to remove e.g. topical information from word embeddings. However, scale isolation may also enable tailored search for particular kinds of information in text or other content, which could enable uses that are beneficial or harmful depending on the use case and whether consent is obtained by relevant parties. The prism layer falls under a general trend of producing more capable neural networks. Such a trend may contribute to increased automation or other changes in labor markets, which may create benefits and harms that depend on the economic and social policies of relevant governing bodies.
\begin{ack}
We would like to thank Jesse Mu, Shikhar Murty, Ben Newman, Mike Wu, Pratyusha Ria Kalluri, and Jesse Michel for useful discussions and comments on drafts. This work was supported in part by DARPA under agreement FA8650-19-C-7923.
\end{ack}

\small

\bibliographystyle{unsrt}
\bibliography{refs}

\begin{thebibliography}{10}

\bibitem{nida1949morphology}
Eugene~A Nida.
\newblock Morphology: The descriptive analysis of words.
\newblock 1949.

\bibitem{cruse1986lexical}
D~Alan Cruse, David~Alan Cruse, D~A Cruse, and D~A Cruse.
\newblock {\em Lexical semantics}.
\newblock Cambridge university press, 1986.

\bibitem{kehler2002coherence}
Andrew Kehler and Andrew Kehler.
\newblock {\em Coherence, reference, and the theory of grammar}.
\newblock CSLI publications Stanford, CA, 2002.

\bibitem{grosz1995centering}
Barbara~J Grosz, Scott Weinstein, and Aravind~K Joshi.
\newblock Centering: A framework for modeling the local coherence of discourse.
\newblock {\em Computational linguistics}, 21(2):203--225, 1995.

\bibitem{thompson1987rhetorical}
Sandra~A Thompson and William~C Mann.
\newblock Rhetorical structure theory: A framework for the analysis of texts.
\newblock {\em IPRA Papers in Pragmatics}, 1(1):79--105, 1987.

\bibitem{hearst1997texttiling}
Marti~A Hearst.
\newblock Texttiling: Segmenting text into multi-paragraph subtopic passages.
\newblock {\em Computational linguistics}, 23(1):33--64, 1997.

\bibitem{blei2003latent}
David~M Blei, Andrew~Y Ng, and Michael~I Jordan.
\newblock Latent dirichlet allocation.
\newblock {\em Journal of machine Learning research}, 3(Jan):993--1022, 2003.

\bibitem{mikolov2013distributed}
Tomas Mikolov, Ilya Sutskever, Kai Chen, Greg~S Corrado, and Jeff Dean.
\newblock Distributed representations of words and phrases and their
  compositionality.
\newblock In {\em Advances in neural information processing systems}, pages
  3111--3119, 2013.

\bibitem{pennington2014glove}
Jeffrey Pennington, Richard Socher, and Christopher~D Manning.
\newblock Glove: Global vectors for word representation.
\newblock In {\em Proceedings of the 2014 conference on empirical methods in
  natural language processing (EMNLP)}, pages 1532--1543, 2014.

\bibitem{kiros2015skip}
Ryan Kiros, Yukun Zhu, Russ~R Salakhutdinov, Richard Zemel, Raquel Urtasun,
  Antonio Torralba, and Sanja Fidler.
\newblock Skip-thought vectors.
\newblock In {\em Advances in neural information processing systems}, pages
  3294--3302, 2015.

\bibitem{hill-etal-2016-learning-distributed}
Felix Hill, Kyunghyun Cho, and Anna Korhonen.
\newblock Learning distributed representations of sentences from unlabelled
  data.
\newblock {\em arXiv preprint arXiv:1602.03483}, 2016.

\bibitem{dai2015document}
Andrew~M Dai, Christopher Olah, and Quoc~V Le.
\newblock Document embedding with paragraph vectors.
\newblock {\em arXiv preprint arXiv:1507.07998}, 2015.

\bibitem{sinoara2019knowledge}
Roberta~A Sinoara, Jose Camacho-Collados, Rafael~G Rossi, Roberto Navigli, and
  Solange~O Rezende.
\newblock Knowledge-enhanced document embeddings for text classification.
\newblock {\em Knowledge-Based Systems}, 163:955--971, 2019.

\bibitem{lin2015hierarchical}
Rui Lin, Shujie Liu, Muyun Yang, Mu~Li, Ming Zhou, and Sheng Li.
\newblock Hierarchical recurrent neural network for document modeling.
\newblock In {\em Proceedings of the 2015 Conference on Empirical Methods in
  Natural Language Processing}, pages 899--907, 2015.

\bibitem{li2015hierarchical}
Jiwei Li, Minh-Thang Luong, and Dan Jurafsky.
\newblock A hierarchical neural autoencoder for paragraphs and documents.
\newblock {\em arXiv preprint arXiv:1506.01057}, 2015.

\bibitem{yang2016hierarchical}
Zichao Yang, Diyi Yang, Chris Dyer, Xiaodong He, Alex Smola, and Eduard Hovy.
\newblock Hierarchical attention networks for document classification.
\newblock In {\em Proceedings of the 2016 conference of the North American
  chapter of the association for computational linguistics: human language
  technologies}, pages 1480--1489, 2016.

\bibitem{tai2015improved}
Kai~Sheng Tai, Richard Socher, and Christopher~D Manning.
\newblock Improved semantic representations from tree-structured long
  short-term memory networks.
\newblock {\em arXiv preprint arXiv:1503.00075}, 2015.

\bibitem{bowman2016fast}
Samuel~R Bowman, Jon Gauthier, Abhinav Rastogi, Raghav Gupta, Christopher~D
  Manning, and Christopher Potts.
\newblock A fast unified model for parsing and sentence understanding.
\newblock {\em arXiv preprint arXiv:1603.06021}, 2016.

\bibitem{ji2014representation}
Yangfeng Ji and Jacob Eisenstein.
\newblock Representation learning for text-level discourse parsing.
\newblock In {\em Proceedings of the 52nd Annual Meeting of the Association for
  Computational Linguistics (Volume 1: Long Papers)}, pages 13--24, 2014.

\bibitem{reimers2019sentence}
Nils Reimers and Iryna Gurevych.
\newblock Sentence-bert: Sentence embeddings using siamese bert-networks.
\newblock {\em arXiv preprint arXiv:1908.10084}, 2019.

\bibitem{conneau2018you}
Alexis Conneau, Germ{\'a}n Kruszewski, Guillaume Lample, Lo{\"\i}c Barrault,
  and Marco Baroni.
\newblock What you can cram into a single vector: Probing sentence embeddings
  for linguistic properties.
\newblock {\em arXiv preprint arXiv:1805.01070}, 2018.

\bibitem{tenney2019bert}
Ian Tenney, Dipanjan Das, and Ellie Pavlick.
\newblock Bert rediscovers the classical nlp pipeline.
\newblock {\em arXiv preprint arXiv:1905.05950}, 2019.

\bibitem{liu2019linguistic}
Nelson~F Liu, Matt Gardner, Yonatan Belinkov, Matthew Peters, and Noah~A Smith.
\newblock Linguistic knowledge and transferability of contextual
  representations.
\newblock {\em arXiv preprint arXiv:1903.08855}, 2019.

\bibitem{hewitt2019structural}
John Hewitt and Christopher~D Manning.
\newblock A structural probe for finding syntax in word representations.
\newblock In {\em Proceedings of the 2019 Conference of the North American
  Chapter of the Association for Computational Linguistics: Human Language
  Technologies, Volume 1 (Long and Short Papers)}, pages 4129--4138, 2019.

\bibitem{oppenheim1999discrete}
Alan~V Oppenheim.
\newblock {\em Discrete-time signal processing}.
\newblock Pearson Education India, 1999.

\bibitem{devlin2018bert}
Jacob Devlin, Ming-Wei Chang, Kenton Lee, and Kristina Toutanova.
\newblock Bert: Pre-training of deep bidirectional transformers for language
  understanding.
\newblock {\em arXiv preprint arXiv:1810.04805}, 2018.

\bibitem{rao2014discrete}
K~Ramamohan Rao and Ping Yip.
\newblock {\em Discrete cosine transform: algorithms, advantages,
  applications}.
\newblock Academic press, 2014.

\bibitem{zuo2014boundary}
Chao Zuo, Qian Chen, and Anand Asundi.
\newblock Boundary-artifact-free phase retrieval with the transport of
  intensity equation: fast solution with use of discrete cosine transform.
\newblock {\em Optics express}, 22(8):9220--9244, 2014.

\bibitem{bovik2009essential}
Alan~C Bovik.
\newblock {\em The essential guide to video processing}.
\newblock Academic Press, 2009.

\bibitem{ahmed1974discrete}
Nasir Ahmed, T\_ Natarajan, and Kamisetty~R Rao.
\newblock Discrete cosine transform.
\newblock {\em IEEE transactions on Computers}, 100(1):90--93, 1974.

\bibitem{butterworth1930theory}
Stephen Butterworth et~al.
\newblock On the theory of filter amplifiers.
\newblock {\em Wireless Engineer}, 7(6):536--541, 1930.

\bibitem{takahasi1951ladder}
H~Takahasi.
\newblock On the ladder-type filter network with tchebysheff response.
\newblock {\em J. Inst. Elec. Commun. Engrs. Japan}, 34(2):65--74, 1951.

\bibitem{mallat1999wavelet}
St{\'e}phane Mallat.
\newblock {\em A wavelet tour of signal processing}.
\newblock Elsevier, 1999.

\bibitem{vaswani2017attention}
Ashish Vaswani, Noam Shazeer, Niki Parmar, Jakob Uszkoreit, Llion Jones,
  Aidan~N Gomez, {\L}ukasz Kaiser, and Illia Polosukhin.
\newblock Attention is all you need.
\newblock In {\em Advances in neural information processing systems}, pages
  5998--6008, 2017.

\bibitem{radford2019language}
Alec Radford, Jeffrey Wu, Rewon Child, David Luan, Dario Amodei, and Ilya
  Sutskever.
\newblock Language models are unsupervised multitask learners.
\newblock {\em OpenAI Blog}, 1(8):9, 2019.

\bibitem{hochreiter1997long}
Sepp Hochreiter and J{\"u}rgen Schmidhuber.
\newblock Long short-term memory.
\newblock {\em Neural computation}, 9(8):1735--1780, 1997.

\bibitem{peters2018deep}
Matthew~E Peters, Mark Neumann, Mohit Iyyer, Matt Gardner, Christopher Clark,
  Kenton Lee, and Luke Zettlemoyer.
\newblock Deep contextualized word representations.
\newblock {\em arXiv preprint arXiv:1802.05365}, 2018.

\bibitem{tenney2019you}
Ian Tenney, Patrick Xia, Berlin Chen, Alex Wang, Adam Poliak, R~Thomas McCoy,
  Najoung Kim, Benjamin Van~Durme, Samuel~R Bowman, Dipanjan Das, et~al.
\newblock What do you learn from context? probing for sentence structure in
  contextualized word representations.
\newblock {\em arXiv preprint arXiv:1905.06316}, 2019.

\bibitem{alain2016understanding}
Guillaume Alain and Yoshua Bengio.
\newblock Understanding intermediate layers using linear classifier probes.
\newblock {\em arXiv preprint arXiv:1610.01644}, 2016.

\bibitem{ettinger2016probing}
Allyson Ettinger, Ahmed Elgohary, and Philip Resnik.
\newblock Probing for semantic evidence of composition by means of simple
  classification tasks.
\newblock In {\em Proceedings of the 1st Workshop on Evaluating Vector-Space
  Representations for NLP}, pages 134--139, 2016.

\bibitem{shi2016does}
Xing Shi, Inkit Padhi, and Kevin Knight.
\newblock Does string-based neural mt learn source syntax?
\newblock In {\em Proceedings of the 2016 Conference on Empirical Methods in
  Natural Language Processing}, pages 1526--1534, 2016.

\bibitem{marcus1993building}
Mitchell Marcus, Beatrice Santorini, and Mary~Ann Marcinkiewicz.
\newblock Building a large annotated corpus of english: The penn treebank.
\newblock 1993.

\bibitem{Jurafsky-etal:1997}
Daniel Jurafsky, Elizabeth Shriberg, and Debra Biasca.
\newblock Switchboard {SWBD}-{DAMSL} shallow-discourse-function annotation
  coders manual, draft 13.
\newblock Technical Report 97-02, University of Colorado, Boulder Institute of
  Cognitive Science, Boulder, CO, 1997.

\bibitem{Shriberg-etal:1998}
Elizabeth Shriberg, Rebecca Bates, Paul Taylor, Andreas Stolcke, Daniel
  Jurafsky, Klaus Ries, Noah Coccaro, Rachel Martin, Marie Meteer, and Carol
  Van Ess-Dykema.
\newblock Can prosody aid the automatic classification of dialog acts in
  conversational speech?
\newblock {\em Language and Speech}, 41(3--4):439--487, 1998.

\bibitem{Stolcke-etal:2000}
Andreas Stolcke, Klaus Ries, Noah Coccaro, Elizabeth Shriberg, Rebecca Bates,
  Daniel Jurafsky, Paul Taylor, Rachel Martin, Marie Meteer, and Carol Van
  Ess-Dykema.
\newblock Dialogue act modeling for automatic tagging and recognition of
  conversational speech.
\newblock {\em Computational Linguistics}, 26(3):339--371, 2000.

\bibitem{lang1995newsweeder}
Ken Lang.
\newblock Newsweeder: Learning to filter netnews.
\newblock In {\em Machine Learning Proceedings 1995}, pages 331--339. Elsevier,
  1995.

\bibitem{kingma2014adam}
Diederik~P Kingma and Jimmy Ba.
\newblock Adam: A method for stochastic optimization.
\newblock {\em arXiv preprint arXiv:1412.6980}, 2014.

\bibitem{merity2016pointer}
Stephen Merity, Caiming Xiong, James Bradbury, and Richard Socher.
\newblock Pointer sentinel mixture models.
\newblock {\em arXiv preprint arXiv:1609.07843}, 2016.

\bibitem{trinh2018learning}
Trieu~H Trinh, Andrew~M Dai, Minh-Thang Luong, and Quoc~V Le.
\newblock Learning longer-term dependencies in rnns with auxiliary losses.
\newblock {\em arXiv preprint arXiv:1803.00144}, 2018.

\bibitem{ebeling1995long}
Werner Ebeling and Alexander Neiman.
\newblock Long-range correlations between letters and sentences in texts.
\newblock {\em Physica A: Statistical Mechanics and its Applications},
  215(3):233--241, 1995.

\bibitem{pavlov2001scaling}
Alexey~N Pavlov, Werner Ebeling, Lutz Molgedey, Amir~R Ziganshin, and Vadim~S
  Anishchenko.
\newblock Scaling features of texts, images and time series.
\newblock {\em Physica A: Statistical Mechanics and its Applications},
  300(1-2):310--324, 2001.

\bibitem{ebeling1994entropy}
Werner Ebeling and Thorsten P{\"o}schel.
\newblock Entropy and long-range correlations in literary english.
\newblock {\em EPL (Europhysics Letters)}, 26(4):241, 1994.

\bibitem{coifman2006diffusion}
Ronald~R Coifman and Mauro Maggioni.
\newblock Diffusion wavelets.
\newblock {\em Applied and Computational Harmonic Analysis}, 21(1):53--94,
  2006.

\bibitem{wang2009multiscale}
Chang Wang and Sridhar Mahadevan.
\newblock Multiscale analysis of document corpora based on diffusion models.
\newblock In {\em Twenty-First International Joint Conference on Artificial
  Intelligence}, 2009.

\bibitem{jain2014short}
Vidit Jain and Jay Mahadeokar.
\newblock Short-text representation using diffusion wavelets.
\newblock In {\em Proceedings of the 23rd International Conference on World
  Wide Web}, pages 301--302, 2014.

\bibitem{sarkar2019scale}
Aakash Sarkar and Marc Howard.
\newblock Scale-dependent relationships in natural language.
\newblock {\em arXiv preprint arXiv:1912.07506}, 2019.

\bibitem{hochreiter2001gradient}
Sepp Hochreiter, Yoshua Bengio, Paolo Frasconi, J{\"u}rgen Schmidhuber, et~al.
\newblock Gradient flow in recurrent nets: the difficulty of learning long-term
  dependencies, 2001.

\bibitem{khandelwal2018sharp}
Urvashi Khandelwal, He~He, Peng Qi, and Dan Jurafsky.
\newblock Sharp nearby, fuzzy far away: How neural language models use context.
\newblock {\em arXiv preprint arXiv:1805.04623}, 2018.

\bibitem{clark2019does}
Kevin Clark, Urvashi Khandelwal, Omer Levy, and Christopher~D Manning.
\newblock What does bert look at? an analysis of bert's attention.
\newblock {\em arXiv preprint arXiv:1906.04341}, 2019.

\bibitem{koutnik2014clockwork}
Jan Koutnik, Klaus Greff, Faustino Gomez, and Juergen Schmidhuber.
\newblock A clockwork rnn.
\newblock {\em arXiv preprint arXiv:1402.3511}, 2014.

\bibitem{el1996hierarchical}
Salah El~Hihi and Yoshua Bengio.
\newblock Hierarchical recurrent neural networks for long-term dependencies.
\newblock In {\em Advances in neural information processing systems}, pages
  493--499, 1996.

\bibitem{chung2016hierarchical}
Junyoung Chung, Sungjin Ahn, and Yoshua Bengio.
\newblock Hierarchical multiscale recurrent neural networks.
\newblock {\em arXiv preprint arXiv:1609.01704}, 2016.

\bibitem{yue2018residual}
Boxuan Yue, Junwei Fu, and Jun Liang.
\newblock Residual recurrent neural networks for learning sequential
  representations.
\newblock {\em Information}, 9(3):56, 2018.

\bibitem{chang2019antisymmetricrnn}
Bo~Chang, Minmin Chen, Eldad Haber, and Ed~H Chi.
\newblock Antisymmetricrnn: A dynamical system view on recurrent neural
  networks.
\newblock {\em arXiv preprint arXiv:1902.09689}, 2019.

\bibitem{arora2016simple}
Sanjeev Arora, Yingyu Liang, and Tengyu Ma.
\newblock A simple but tough-to-beat baseline for sentence embeddings.
\newblock 2016.

\bibitem{zhang2000forenet}
Y~Zhang and Lai-Wan Chan.
\newblock Forenet: fourier recurrent networks for time series prediction.
\newblock 2000.

\bibitem{zhang2018learning}
Jiong Zhang, Yibo Lin, Zhao Song, and Inderjit~S Dhillon.
\newblock Learning long term dependencies via fourier recurrent units.
\newblock {\em arXiv preprint arXiv:1803.06585}, 2018.

\bibitem{shen2018ordered}
Yikang Shen, Shawn Tan, Alessandro Sordoni, and Aaron Courville.
\newblock Ordered neurons: Integrating tree structures into recurrent neural
  networks.
\newblock {\em arXiv preprint arXiv:1810.09536}, 2018.

\bibitem{abdel2012applying}
Ossama Abdel-Hamid, Abdel-rahman Mohamed, Hui Jiang, and Gerald Penn.
\newblock Applying convolutional neural networks concepts to hybrid nn-hmm
  model for speech recognition.
\newblock In {\em 2012 IEEE international conference on Acoustics, speech and
  signal processing (ICASSP)}, pages 4277--4280. IEEE, 2012.

\bibitem{hinton2012deep}
Geoffrey Hinton, Li~Deng, Dong Yu, George~E Dahl, Abdel-rahman Mohamed, Navdeep
  Jaitly, Andrew Senior, Vincent Vanhoucke, Patrick Nguyen, Tara~N Sainath,
  et~al.
\newblock Deep neural networks for acoustic modeling in speech recognition: The
  shared views of four research groups.
\newblock {\em IEEE Signal processing magazine}, 29(6):82--97, 2012.

\bibitem{nussbaumer1981fast}
Henri~J Nussbaumer.
\newblock The fast fourier transform.
\newblock In {\em Fast Fourier Transform and Convolution Algorithms}, pages
  80--111. Springer, 1981.

\bibitem{mathieu2013fast}
Michael Mathieu, Mikael Henaff, and Yann LeCun.
\newblock Fast training of convolutional networks through ffts.
\newblock {\em arXiv preprint arXiv:1312.5851}, 2013.

\bibitem{khan2017scene}
Salman~H Khan, Munawar Hayat, and Fatih Porikli.
\newblock Scene categorization with spectral features.
\newblock In {\em Proceedings of the IEEE International Conference on Computer
  Vision}, pages 5638--5648, 2017.

\bibitem{cheng2015exploration}
Yu~Cheng, Felix~X Yu, Rogerio~S Feris, Sanjiv Kumar, Alok Choudhary, and Shi-Fu
  Chang.
\newblock An exploration of parameter redundancy in deep networks with
  circulant projections.
\newblock In {\em Proceedings of the IEEE International Conference on Computer
  Vision}, pages 2857--2865, 2015.

\bibitem{karras2019style}
Tero Karras, Samuli Laine, and Timo Aila.
\newblock A style-based generator architecture for generative adversarial
  networks.
\newblock In {\em Proceedings of the IEEE Conference on Computer Vision and
  Pattern Recognition}, pages 4401--4410, 2019.

\bibitem{rippel2015spectral}
Oren Rippel, Jasper Snoek, and Ryan~P Adams.
\newblock Spectral representations for convolutional neural networks.
\newblock In C.~Cortes, N.~D. Lawrence, D.~D. Lee, M.~Sugiyama, and R.~Garnett,
  editors, {\em Advances in Neural Information Processing Systems 28}, pages
  2449--2457. Curran Associates, Inc., 2015.

\bibitem{khan2019regularization}
Salman~H Khan, Munawar Hayat, and Fatih Porikli.
\newblock Regularization of deep neural networks with spectral dropout.
\newblock {\em Neural Networks}, 110:82--90, 2019.

\bibitem{wu2016google}
Yonghui Wu, Mike Schuster, Zhifeng Chen, Quoc~V Le, Mohammad Norouzi, Wolfgang
  Macherey, Maxim Krikun, Yuan Cao, Qin Gao, Klaus Macherey, et~al.
\newblock Google's neural machine translation system: Bridging the gap between
  human and machine translation.
\newblock {\em arXiv preprint arXiv:1609.08144}, 2016.

\bibitem{wolf2019transformers}
Thomas Wolf, Lysandre Debut, Victor Sanh, Julien Chaumond, Clement Delangue,
  Anthony Moi, Pierric Cistac, Tim Rault, R{\'e}mi Louf, Morgan Funtowicz,
  et~al.
\newblock Transformers: State-of-the-art natural language processing.
\newblock {\em arXiv preprint arXiv:1910.03771}, 2019.

\bibitem{chen1977fast}
Wen-Hsiung Chen, CH~Smith, and SC~Fralick.
\newblock A fast computational algorithm for the discrete cosine transform.
\newblock {\em IEEE Transactions on communications}, 25(9):1004--1009, 1977.

\end{thebibliography}

\newpage
\appendix
\section{Spectral Band Allocation}

We have $n$ frequencies we wish to allocate into $k$ bands such that $\sum_i a_i = n$, where $a_i$ denotes the number of frequencies allocated to band $i$.

We begin by allocating each band one frequency, leaving $(n - k)$ frequencies remaining. Next, we choose a base $b$ (we use $b=4$) and generate allocation scores $s_i = b^i$ which we normalize into fractional allocations $\tilde a_i =  (n-k) b^i / \sum_i (s_i) $. However, allocations must be whole numbers, so we produce conservative allocations $ \check a_i = \lfloor \tilde a_i \rfloor$, and then allocate the remaining frequencies in descending order of $\tilde a_i - \check a_i$ (i.e. whichever bands were closest to receiving another frequency) until $\sum_i a_i = n$. 

However, we note again that in practice, one might wish to smoothly vary the endpoints of the spectral band between neurons, rather than choosing only 5 canonical bands. In addition, one could specifically choose bands for a task based on their corresponding periods to include or exclude particular scales of interest.

\section{Additional Ablations}

We present additional ablations here, with new results displayed in \textbf{bold}. All results are averages of three trials.

\subsection{Individual sectors of the BERT + prism model}

What information is accessible from individual sectors of the BERT + prism model? For the lowest frequency sector of the BERT + prism model, topic classification probing accuracy is \textbf{45.1\%}, versus \textbf{5.3\%} for the highest-frequency sector and 51.0\% for the full model. For the highest-frequency sector, POS tagging probing accuracy is \textbf{84.1\%}, versus \textbf{16.8\%} for the lowest-frequency sector and 94.4\% for the full model. This suggests that the \textsc{high} and \textsc{low} frequency bands are largely but not entirely responsible for the BERT + prism model's performance on POS tagging and topic classification, respectively.

\subsection{Performance of BERT representations without finetuning on WikiText-103}

To confirm that the additional pretraining on WikiText-103 does not harm BERT, producing an artificially weak baseline, we compare probing performance on the original pretrained BERT model. The original model achieves an accuracy of \textbf{94.6\%} for POS tagging, \textbf{41.8\%} for dialog acts, and \textbf{28.9\%} for topic classification, slightly worse than the better model that was trained longer on WikiText-103 (95.9\%, 47.1\%, 32.2\% respectively).

\subsection{Placing prism layers after every BERT layer}

Can we obtain better representations by adding a prism layer after \emph{each} BERT layer instead of just the last? We find that this model  produces worse representations than BERT + prism, achieving \textbf{45.2\%} accuracy on topic classification (-5.8\%), \textbf{51.8\%} on dialog speech acts (-2.2\%) and \textbf{94.0\%} on POS tagging (-0.4\%). We suspect this may be because the removal of spectral information deeper in the network reduces the model's effective capacity, interfering with its ability to produce useful representations.

\section{Data and preprocessing}

We tokenize all data inputs using a WordPiece tokenizer \cite{wu2016google} from an external library \cite{wolf2019transformers}. For each gold label (e.g., a part of speech tag, a dialog speech act annotation, or a topic), we then perform our probing experiments on each representation from the resulting $N$ tokens, weighting the loss and accuracy for each embedding's prediction by $1 / |N|$.

For dialog speech acts and masked language modeling, which have long inputs, we chunk each input into segments of at most length 510, then append the two special tokens before feeding them into our models. For part of speech tagging and topic classification, we discard excess tokens. We use the traditional train/validation/test splits for all models:

\begin{itemize}
    \item The 20 Newsgroups dataset has approximately 20k documents, with 60\% for training and the remainder for testing. \url{http://qwone.com/~jason/20Newsgroups/}
    \item The Switchboard Dialog Speech Acts corpus contains around 1.1k training transcripts and 19 validation transcripts. \url{https://github.com/cgpotts/swda}
    \item The Penn Treebank has approximately 38.2k examples for training and 5.5k for validation. \url{https://catalog.ldc.upenn.edu/LDC99T42}
    \item Wikitext 103 has around 500MB of text for training and 1.1MB for validation. \url{https://blog.einstein.ai/the-wikitext-long-term-dependency-language-modeling-dataset/}
\end{itemize}

\section{Complexity}

The DCT can be computed efficiently using the FFT in $O(N \log N)$ \cite{chen1977fast}.

\section{Computational Time and Resources}

All experiments were performed on single Titan XP GPUs. Each experiment took approximately one day, while MLM training with the prism layer took approximately 8 hours. Collecting losses for Figure \ref{fig:cliff} took on the order of minutes.

\end{document}